\documentclass[letterpaper, 10 pt, conference]{ieeeconf} 

\IEEEoverridecommandlockouts
\usepackage{cite}
\usepackage{soul}
\usepackage{xcolor}
\usepackage{caption}
\usepackage{amsmath,amssymb,amsfonts}
\usepackage{graphicx}
\usepackage{float}
\usepackage{textcomp}
\usepackage{url}
\usepackage{booktabs}
\usepackage[hypertexnames=false]{hyperref}
\usepackage{multirow}
\usepackage{array}
\usepackage{tabularx}
\usepackage{algorithm}
\usepackage{algorithmic}
\usepackage[acronym]{glossaries}

\glsdisablehyper
\graphicspath{{figs/}}

\begin{document}

\title{\LARGE \bf Calf-Integrated Arms for Bimanual Quadruped Loco-Manipulation}

\author{Yan Pan\textsuperscript{1}, Yuanchuan Ren\textsuperscript{1}, Chipui Chan\textsuperscript{1}, Jingcheng Sun\textsuperscript{1} and Chengxu Zhou\textsuperscript{1}
\thanks{This work was partially supported by the Advanced Research and Invention Agency [grant number SMRB-SE01-P06]. 
}
\thanks{\textsuperscript{1}Dept. of Computer Science, University College London, UK. {\tt\small chengxu.zhou@ucl.ac.uk}}
}

\maketitle

\begin{abstract}
Most quadruped loco-manipulation designs trade manipulation capability against stance. A trunk-mounted arm sits high and usually carries a single arm; using the legs as manipulators lifts the manipulating leg off the ground; and even leg-mounted grippers reach two-handed tasks only by rearing onto the hind legs. This paper integrates a manipulator with a prismatic slider, two revolute joints, and a gripper into each front calf of a Unitree Go2. The two arms grasp objects at ground level and manipulate with both hands while all four feet stay planted, without rearing. With one arm carrying, the base stays free to walk. A vision-language model sequences skills from a predefined library at each skill boundary, conditioned on the head-camera image and task state, for long-horizon autonomy. In simulation, the design performs three bimanual tasks: a long-horizon cabinet task under autonomous skill selection, a cooperative two-handed lift, and an inter-arm handover.
\end{abstract}


\section{Introduction}
\label{sec:introduction}

Quadruped robots move capably through cluttered, human-scale environments~\cite{joey2025adapt}, yet endowing them with manipulation, the ability to act on the objects around them, remains an open design problem. Loco-manipulation, the combination of locomotion and manipulation, would extend them to tasks such as search and rescue, inspection, and logistics. Consider clearing a cluttered room: the robot must pick litter off the floor, carry a basket in one hand while opening a cabinet with the other, and choose the order of these actions itself. This calls for three capabilities at once: reaching objects at ground level, using both hands together, and sequencing skills over a long horizon.

Existing loco-manipulation designs rarely offer all three (Section~\ref{sec:related}). A trunk-mounted arm reaches objects but sits high on the body and usually carries a single arm. Using the bare legs as manipulators gives up the support stance, and leg-mounted grippers free both hands only by rearing onto the hind legs, which pins the base in place. No design keeps all four feet planted while manipulating with two hands down to ground level, nor pairs this with a planner that selects skills from the current scene and task state. The integrated leg-arm design proposed here provides all three capabilities: a manipulator built into each front calf of a Unitree Go2, so the quadruped grasps objects at ground level with both arms.

This paper makes two contributions:
\begin{enumerate}
    \item Integrating a manipulator with a prismatic slider, two revolute joints, and a gripper into each front calf of a Unitree Go2, enabling bimanual loco-manipulation and ground-level grasping.
    \item Introducing a vision-language model that selects the next skill from a predefined library at each skill boundary, conditioned on the head-camera image and task state, demonstrated on a long-horizon bimanual task in simulation.
\end{enumerate}

The paper details the mechanical design (Section~\ref{sec:mech_design}) and the control framework (Section~\ref{sec:algorithms}), reports the simulation study including the observed failure modes (Section~\ref{sec:experiment_result}), and concludes in Section~\ref{sec:conclusion}.

\section{Related Work}
\label{sec:related}

Loco-manipulation designs differ in where the manipulator is placed. The most common approach mounts a dedicated arm on the trunk: articulated arms on ANYmal~\cite{bellicoso2019alma} and Spot~\cite{zimmermann2021go} perform dynamic grasping, model-predictive and whole-body controllers coordinate the arm with the base~\cite{sambhus2025nonlinear,li2022whole}, and learning-based controllers extend this to manipulation policies~\cite{ha2024umi} and to opening and traversing doors~\cite{zhang2024learning}. The arm reaches a large workspace, but it sits high on the body, so reaching ground-level objects means reaching down past the legs, and the torso typically carries a single arm, which rules out simultaneous two-handed tasks.

A second line reuses the legs. Reinforcement-learning policies turn a bare leg into a button-presser~\cite{cheng2023legs} or a pusher~\cite{arm2024pedipulate}, multi-contact optimisation manipulates a ball with the legs~\cite{yang2020dynamic}, and learned interlimb coordination broadens the loco-manipulation repertoire~\cite{zhu2025versatile}. A bare foot, however, offers little prehension, and the manipulating leg must leave the support polygon. Closest to our design, LocoMan already mounts compact grippers on the front calves~\cite{lin2024locoman}; it reaches two-handed tasks by rearing onto the hind legs, freeing both front manipulators over a large workspace but lifting the front feet and fixing the base.

\begin{table}[t]
    \caption{Positioning of the proposed design against existing loco-manipulation paradigms. \emph{Simple hardware modification}: a small bolt-on change such as a foot-mounted gripper, rather than replacing the calf. \emph{Bimanual loco-manipulation}: two-handed manipulation while the base remains free to walk.}
    \label{tab:positioning}
    \centering
    \setlength{\tabcolsep}{2pt}
    \begin{tabular}{@{}lcccc@{}}
        \toprule
        \textbf{Feature} & \textbf{Trunk-arm} & \textbf{Leg-only} & \textbf{LocoMan} & \textbf{Ours} \\
        \midrule
        Prehensile ground reach                 & \checkmark & --         & \checkmark & \checkmark \\
        Simple hardware modification        & --         & \checkmark & \checkmark & --         \\
        Four-foot stance in manipulation & \checkmark & --         & --         & \checkmark \\
        Bimanual loco-manipulation       & --         & --         & --         & \checkmark \\
        \bottomrule
    \end{tabular}
\end{table}

A third line reconfigures the legs themselves: embedded shape morphing~\cite{sun2023embedded}, sprawling-angle adaptation~\cite{yuan2024design}, modular legs~\cite{kim2017snapbot}, telescopic legs~\cite{mohamed2024design}, and tensegrity or flexible limbs~\cite{cui2022design,huang2021quadruped}. These designs primarily target locomotion versatility and terrain adaptability rather than prehensile manipulation.

The design we propose here also integrates the manipulator into each front calf, but adds a yaw joint~($q_3$) that sweeps both grippers to the body centreline; the two front arms then cooperate on a single object without rearing onto the hind legs, with all four feet in stance and the base free to walk (Table~\ref{tab:positioning}), trading a smaller bimanual workspace than rearing affords for stability and a walk-ready base.

Beyond hardware, a recent line drives legged loco-manipulation from a language model rather than a fixed script. Large language models sequence locomotion and manipulation skills for long-horizon quadruped tasks~\cite{ouyang2024longhorizon}, and grounded language planners build hierarchical task graphs over motion primitives for autonomous loco-manipulation~\cite{wang2024hypermotion}. The high-level planner used here follows this paradigm, selecting a skill from a fixed library at each step from the camera image and task state; the contribution is not the planner but the leg-integrated hardware that lets such a plan run with all four feet planted and both hands free.

\section{Mechanical Design}
\label{sec:mech_design}
The design integrates a manipulator arm with a sliding gripper into each front leg of a Unitree Go2. Placing the gripper on the leg rather than high on the trunk lets it reach ground-level objects and retract while walking, so each front leg both walks and manipulates. The two mirror-identical arms together enable bimanual loco-manipulation without lifting a leg off the ground.

\begin{figure}[t]
    \centering
    \includegraphics[width=0.99\columnwidth]{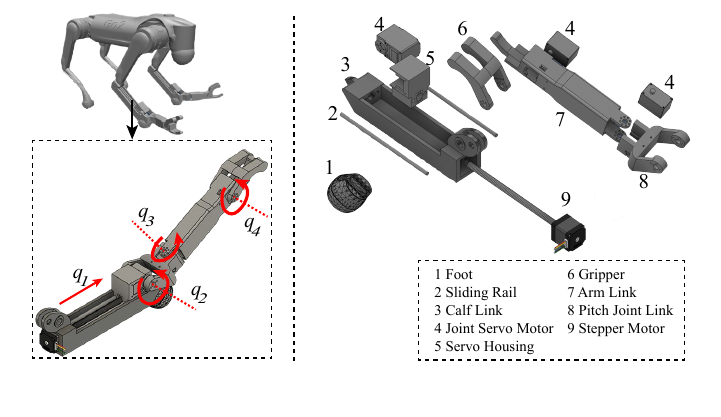}
    \caption{Design of the integrated leg-arm. \emph{Left}:~joint configuration, with the added joints $q_1$ (prismatic slider), $q_2$ (pitch), $q_3$ (yaw), and $q_4$ (parallel-jaw gripper) forming the manipulator and red arrows showing each joint's motion. \emph{Right}:~exploded view of one calf module.}
    \label{fig:leg_module}
\end{figure}

\paragraph{Joint configuration}
Figure~\ref{fig:leg_module} shows the front-leg module. Each front leg has seven degrees of freedom (DoF): three original leg joints that position the lower leg and four added joints that form the manipulator. Of the added joints, $q_1$ is a prismatic slider with a $0.105$\,m stroke, housed inside the rebuilt calf rather than bolted on, keeping the structure compact; it slides the gripper down to pick objects off the ground. The revolute joint $q_2$ pitches the gripper for fine approach. The revolute joint $q_3$ rotates about the vertical (yaw) axis, sweeping the gripper towards the body centreline so that the two front arms can come close enough to cooperate on the same object. The parallel-jaw gripper $q_4$ opens to 4\,cm. The slider $q_1$ is driven by a stepper motor with a non-backdrivable drive, which holds a lifted load without continuous power, while the revolute joints $q_2$, $q_3$ and the gripper $q_4$ use servos for closed-loop position control.

\begin{figure}[t]
    \centering
    \includegraphics[width=0.99\columnwidth]{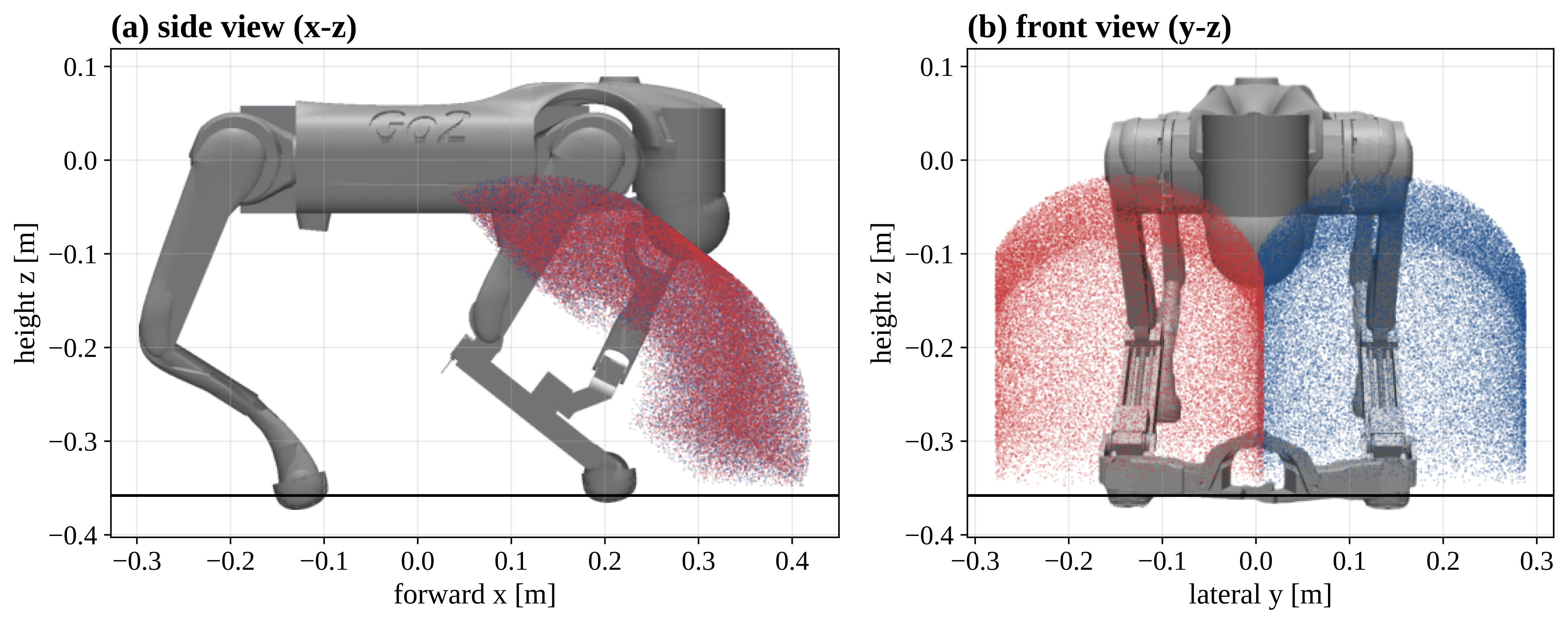}
    \caption{Reachable workspace of the two front-leg grippers: \emph{(a)}~side and \emph{(b)}~front views. The left (blue) and right (red) workspaces reach the ground and overlap at the body centreline, enabling bimanual cooperation.}
    \label{fig:workspace}
\end{figure}

\paragraph{Workspace}
Figure~\ref{fig:workspace} shows the reachable workspace of each gripper, sampled across the manipulator joints $(q_1,q_2,q_3)$ by forward kinematics with the legs in their standing stance. The pinch point reaches from $0.01$\,m above the floor up to $0.34$\,m, the $0.105$\,m slider and the pitch joint $q_2$ together covering this vertical range, while the yaw joint $q_3$ sweeps each gripper towards the body centreline, where the left and right workspaces overlap. Integrating the manipulator into the calf places this reach close to the ground: the arm base sits $0.18$\,m above the floor against $0.36$\,m for a trunk mount, so a trunk-mounted arm would need twice the reach to grasp the same floor-level object.

\paragraph{Locomotion trade-off}
Integrating the manipulator into the calf trades some locomotion margin for the added reach: the rebuilt calf is heavier and bulkier, raising the distal mass and swing inertia. It also stops the knee $8^\circ$ short of full fold, so only the deepest crouch is affected; the leg-lift used in normal walking stays well within range, leaving step clearance during locomotion unchanged.

\section{Control Architecture}
\label{sec:algorithms}


\subsection{Architecture overview}

\begin{figure*}
\centering
\includegraphics[width=0.99\textwidth]{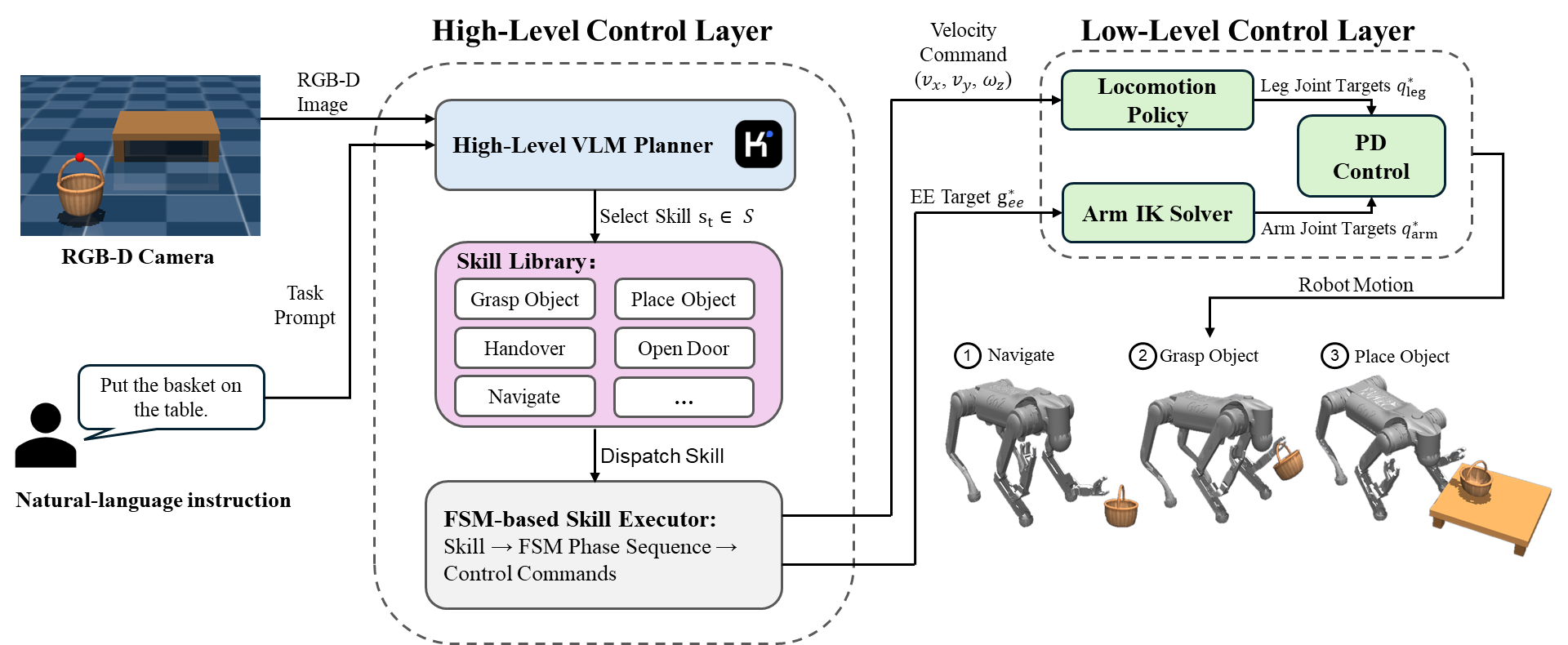}
\caption{Two-layer control architecture. The high-level planner uses the head-mounted RGB-D observation, the user task prompt, and a structured world state to select one skill at a time from the library $\mathcal{S}$. The selected FSM interfaces with the learned locomotion controller for the 12 leg joints and the DLS IK solver for the front-leg arm joints $q_1$--$q_3$, while the gripper joint $q_4$ is commanded directly. The bottom strip illustrates the navigate, grasp, and place phases of a representative task.}
\label{fig:control_arch}
\end{figure*}

Figure~\ref{fig:control_arch} gives an overview of the proposed two-layer control architecture. At the high level, a vision-language model (VLM), Kimi K2.6 served through cloud API, selects discrete skills from a predefined library $\mathcal{S}$. At the low level, each selected skill is executed by a finite state machine (FSM), which sends body-frame velocity commands $\mathbf{v}_{\mathrm{cmd}}=[v_x,v_y,\omega_z]^\top$ to the learned locomotion policy and Cartesian end-effector targets $\mathbf{g}^\star_{\mathrm{ee}}$ to a 3-DoF damped least-squares (DLS) inverse kinematics (IK) solver. The resulting leg and arm joint targets are tracked by position-controlled actuators.

\subsection{Algorithm}

\paragraph{High-level skill planner}
The world state stores task-specific binary progress flags and an execution history of completed skills. Queries run on a background thread so the locomotion policy keeps stance. Each skill runs to completion before the next query, so the VLM is queried once per skill rather than at every control step.

\paragraph{Skill library}
The library $\mathcal{S}$ is a set of skills in two modes, single-arm and bimanual. Single-arm skills act with one front arm while the other is folded; bimanual skills coordinate both front arms. Each skill is a short FSM state sequence, so the planner only chooses which skill to run, not how the joints move. Table~\ref{tab:fsm} summarises the common phases and the extra coordination phases used by the bimanual skills.

\begin{table}
\caption{FSM phases used by the skills in $\mathcal{S}$. Common phases are reused across skills; bimanual tasks add a short coordination block.}
\label{tab:fsm}
\centering
\small
\begin{tabularx}{\columnwidth}{@{}lX@{}}
\toprule
\textbf{State} & \textbf{Behaviour} \\
\midrule
\multicolumn{2}{@{}l}{\emph{Common phases.}} \\
\textsc{Search}       & Yaw in place until the goal pose is locked. \\
\textsc{Approach}     & Walk to a body-relative standoff from the goal. \\
\textsc{Pre-align}    & Rotate the body to the yaw required by the next phase. \\
\textsc{Extend}       & Smoothly move the active arm from the hold pose to the reach pose. \\
\textsc{Servo}        & Track the grasp target with IK while body velocity reduces the residual. \\
\textsc{Close}        & Close $q_4$ and record the target's resting height. \\
\textsc{Lift-test}    & Raise the end effector; success requires object lift of at least $\delta_{\mathrm{lift}}$. \\
\textsc{Recover}      & Back up, reacquire the goal, or retry after vision loss or timeout. \\
\textsc{Done}         & Hold the terminal success state. \\
\midrule
\multicolumn{2}{@{}l}{\emph{Handover phases.}} \\
\textsc{Set-down}     & The sending arm lowers the object to a transfer pose and opens. \\
\textsc{Re-grasp}     & The receiving arm closes on the object. \\
\midrule
\multicolumn{2}{@{}l}{\emph{Cooperative-lift phases.}} \\
\textsc{Dual-align}   & Both arms align with opposite ends of the object. \\
\textsc{Dual-close}   & Both grippers close simultaneously. \\
\textsc{Dual-lift}    & Both sliders raise the object together. \\
\bottomrule
\end{tabularx}
\end{table}

\paragraph{Locomotion module}
Locomotion uses a neural-network policy trained in Isaac Lab~\cite{mittal2025isaaclab} on the modified morphology; the legs are driven by this learned policy and the arm is positioned by IK, splitting contact-rich locomotion from free-space positioning. The 45-dimensional observation is the base angular velocity, the body-frame gravity vector, the velocity command $\mathbf{v}_{\mathrm{cmd}}$, the 12 leg-joint positions and velocities,  and the previous action; the 12-dimensional output is a position residual on a fixed nominal pose, applied at 50\,Hz as the leg-joint targets. With all four feet in stance throughout, the policy retains the unmodified Go2 velocity-tracking interface.

\paragraph{Perception module}
Each graspable object carries a small red marker at its grasp point. RGB thresholding ($R>150$, $G<80$, $B<80$) segments the marker in the image. To locate it in 3D, two opposite points on the marker's outline are back-projected through the depth image into world coordinates, and the grasp point $\mathbf{p}_g$ is taken as their midpoint, which lands on the marker's true centre rather than the nearer camera-facing surface. A detection touching the image border is flagged \emph{partial} and replaced by the average of recent full detections.

\paragraph{Manipulation module}
The grasp places the gripper's pinch point on $\mathbf{p}_g$: a 3-DoF position IK on $(q_1,q_2,q_3)$, while $q_4$ only opens or closes. A DLS solver~\cite{wampler1986manipulator} on the position Jacobian $J_p\in\mathbb{R}^{3\times 3}$ gives the joint increments
\begin{equation}
    \Delta\mathbf{q} = J_p^{\!\top}(J_pJ_p^{\!\top}+\lambda^2 I)^{-1}\,(\mathbf{g}^{\star}_{\mathrm{ee}}-\mathbf{x}_{\mathrm{ee}}(\mathbf{q})),
    \label{eq:dls}
\end{equation}
where $\mathbf{x}_{\mathrm{ee}}(\mathbf{q})$ is the current pinch-point position from forward kinematics, $\mathbf{g}^{\star}_{\mathrm{ee}}$ the Cartesian target, $I$ the $3\times 3$ identity, and $\lambda$ a damping factor that trades tracking accuracy for stability near kinematic singularities. The update is applied with per-iteration step clipping and joint-limit projection, and runs in two modes. While the arm reaches out open-loop, the solver starts from several spread-out configurations and keeps the lowest-residual one, finding a globally good reach. Once it tracks the target closed-loop, it iterates only from the current configuration so the motion stays continuous. During reaching, an optional cap on the slider $q_1$ makes the solver extend the prismatic axis rather than fold the gripper under the body.

\section{Simulation Study}
\label{sec:experiment_result}
\label{sec:experimental_setup}
\label{sec:results}

The design is evaluated entirely in simulation. This section presents three typical bimanual loco-manipulation tasks (Section~\ref{sec:res_demo}), contrasts them with the alternative designs that cannot perform all three (Section~\ref{sec:comparison}), and then traces the vision-language model's reasoning and skill selection on the long-horizon task (Section~\ref{sec:vlm_trace}).

\subsection{Simulation Setup}
\label{sec:setup}

In each task the robot receives a natural-language instruction, such as \emph{place the basket in the cabinet}, and a scene whose object poses are unknown in advance; it must select and execute skills from the library $\mathcal{S}$ to satisfy the instruction, using only the head-mounted RGB-D camera for perception. All tasks require both front-leg arms to coordinate.

\paragraph{Platform}
\label{sec:sim_env}
Training runs in Isaac Lab~\cite{mittal2025isaaclab} (Isaac Sim 5.1.0) on the Go2 with the integrated calf manipulators; evaluation runs in MuJoCo. Training uses 4096 parallel environments at 200\,Hz with decimation factor 4 (a 50\,Hz control loop) and 20\,s episodes, completing in 2\,hours on a single NVIDIA GeForce RTX\,5090.

\paragraph{Training}
\label{sec:config}

The locomotion policy is trained with proximal policy optimisation (PPO) for 5000 iterations, using a $[512,256,128]$ actor and a matching critic, under domain randomisation over friction ($[0.3,1.2]$), restitution ($[0,0.15]$), and added base mass ($[-1,+3]$\,kg). The domain-randomisation ranges and network architecture are the only training hyperparameters material to this paper; all other settings follow standard Isaac Lab locomotion defaults.

\subsection{Versatile Bimanual Loco-Manipulation}
\label{sec:res_demo}
The design is exercised on three bimanual loco-manipulation tasks in simulation, each carried out with all four feet in stance.

\begin{figure*}[t]
    \centering
    \includegraphics[width=0.99\textwidth]{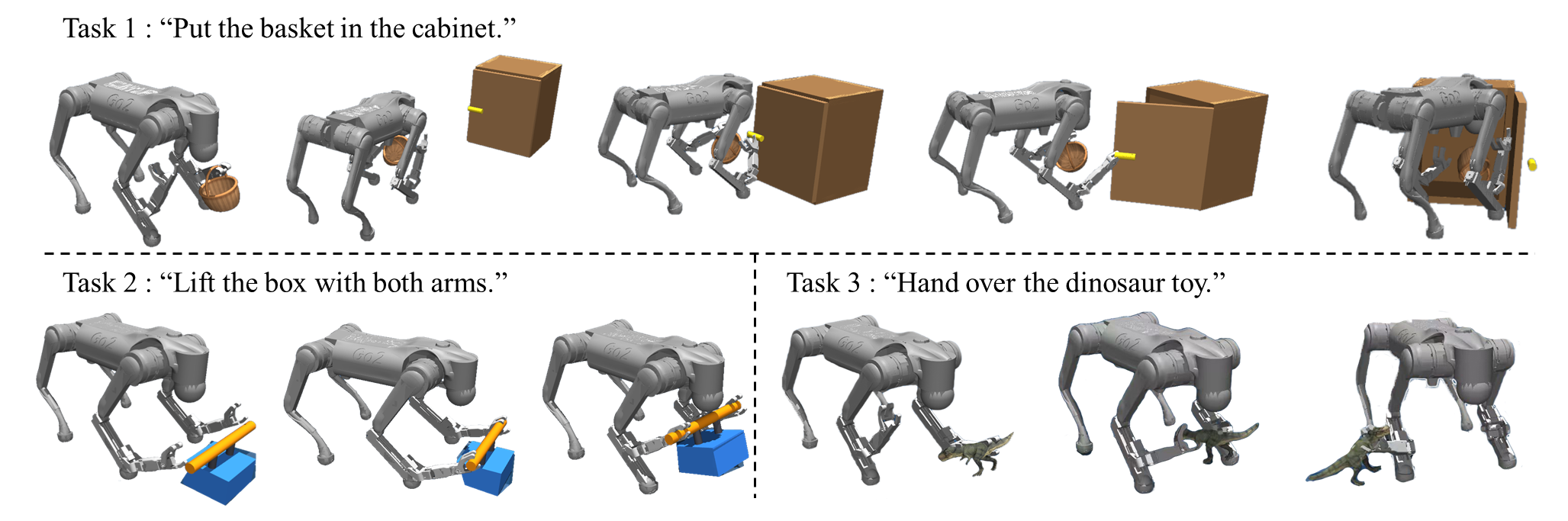}
    \caption{The three bimanual loco-manipulation tasks in simulation, each carried out with all four feet in stance. \emph{Top}: the long-horizon cabinet task, where the left arm carries a basket, the robot walks to the cabinet, the right arm opens the door, and the basket is placed inside. \emph{Bottom left}: the cooperative two-handed lift, where both arms grasp a single box and raise it together. \emph{Bottom right}: the left-to-right handover, where the object passes between grippers at the frontal centreline.}
    \label{fig:demos}
\end{figure*}

\paragraph{Long-horizon cabinet task}
The left arm carries a basket while the robot walks to a cabinet, the right arm opens the door, and the basket is placed inside (Figure~\ref{fig:demos}, top). The task chains locomotion, door opening, and placement; a vision-language model selects the next skill at each stage from the head-camera image and task state.

\paragraph{Cooperative two-handed lift}
Both front arms grasp a single box by its two side handles and raise it together (Figure~\ref{fig:demos}, bottom left). Lifting one object with two grippers needs the left and right reachable sets to overlap at the frontal centreline, which the yaw joints provide.

\paragraph{Left-to-right handover}
The left arm grasps an object, transfers it to the right arm at the centreline, and the right arm sets it down to the side (Figure~\ref{fig:demos}, bottom right). The object passes between grippers without being placed down, again exploiting the centreline workspace overlap.

\subsection{Comparison with alternative designs}
\label{sec:comparison}
The demonstrations highlight what the planted-stance design enables on these tasks. A single trunk-mounted arm~\cite{bellicoso2019alma} cannot carry the basket and open the door at once, so it cannot perform the cabinet task; LocoMan manipulates with both hands by rearing onto the hind legs~\cite{lin2024locoman}, which fixes the base and rules out the repositioning these tasks require. The design here keeps all four feet planted throughout, so the cooperative lift, the handover, and the cabinet approach execute with the base free to walk or hold position.

\subsection{VLM skill selection and reasoning}
\label{sec:vlm_trace}
\begin{figure*}[t]
    \centering
    \includegraphics[width=0.99\textwidth]{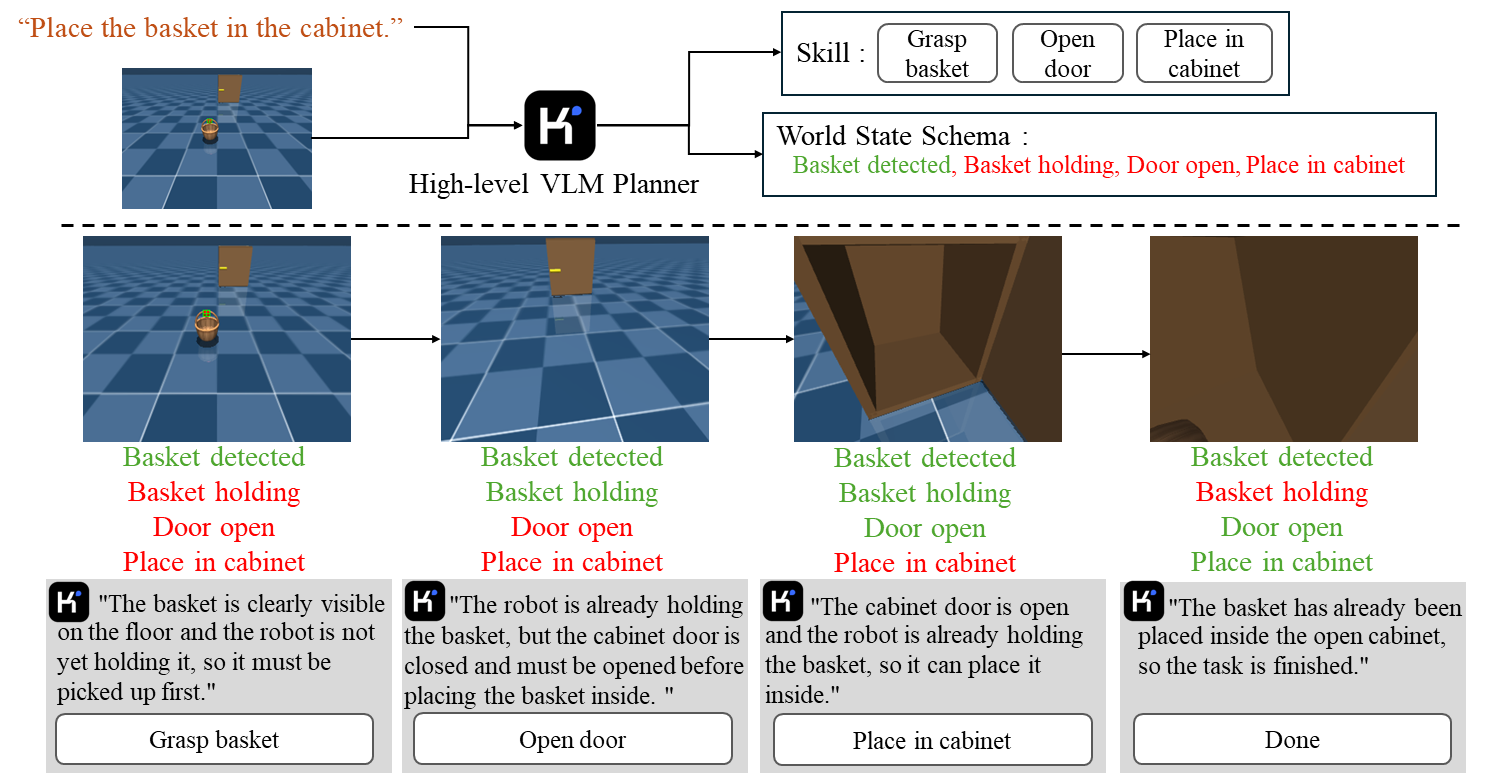}
    \caption{Vision-language planner on the cabinet task. From the instruction and the head-camera image (top), the planner sets up the world-state flags and the skills for the task. At each step (bottom) it reads the image, evaluates the flags, gives its reasoning, and selects the next skill: grasp basket, open door, place in cabinet, and finally done. Green flags are satisfied, red are not.}
    \label{fig:vlm_trace}
\end{figure*}

To show how the planner reasons, consider one run of the cabinet task, driven by the single instruction \emph{“Place the basket in the cabinet.”} with no scripted skill order (Figure~\ref{fig:vlm_trace}). From the instruction the planner sets up a world-state schema, the binary flags that track progress (basket detected, basket held, door open, basket placed). It then loops: at each step it reads the head-camera image, updates the flags, and selects one skill together with its reasoning. The order is not fixed but emerges from the flags, here grasp, open door, place, and done; the planner would reorder or repeat skills if the scene required it, such as reopening a door found shut. One instruction thus drives the task end to end, the autonomous long-horizon behaviour the design targets.

Across the whole task the planner is queried only four times, once per skill boundary. Robot motion is fast, 18.6\,s in total: 7.1\,s to grasp the basket, 9.1\,s to walk to the cabinet and pull the door open, and 2.4\,s to place it inside. The remaining time is planner latency, 16--40\,s per query while the robot holds a stable stance and the hosted Kimi model returns the next skill.

\subsection{Robustness to depth-sensor noise}
\label{sec:perception}
We test the marker-based grasp perception under depth noise. Injecting zero-mean Gaussian noise into the depth measurements (8 fixed basket placements, each repeated under 20 independent noise samples, for 160 trials per level), grasp success stays at 70--74\% up to 10\,mm standard deviation and falls to 54\% at 30\,mm, because the closed-loop controller re-estimates $\mathbf{p}_g$ every frame and zero-mean errors largely cancel. Commodity RGB-D sensors exhibit only millimetre-scale depth noise at these distances, well within the range the pipeline tolerates.

\subsection{Discussion on Limitations}
\label{sec:failure_modes}

Mechanically, the arm has no roll joint, so the gripper cannot rotate about its approach axis. Objects whose handles require this rotation, such as a kettle, cannot be grasped without adding a roll joint.

The skill library has finite coverage. The VLM only selects among predefined skills and creates no new skills or states, so a prompt it understands semantically is still infeasible if it requires a behaviour the library does not provide. Tasks such as re-grasping at a new contact point, rotating an object in hand, avoiding obstacles while carrying, or changing strategy after a failed grasp therefore cannot be solved by prompt reasoning alone.

\begin{figure}[t]
    \centering
    \includegraphics[width=0.99\columnwidth]{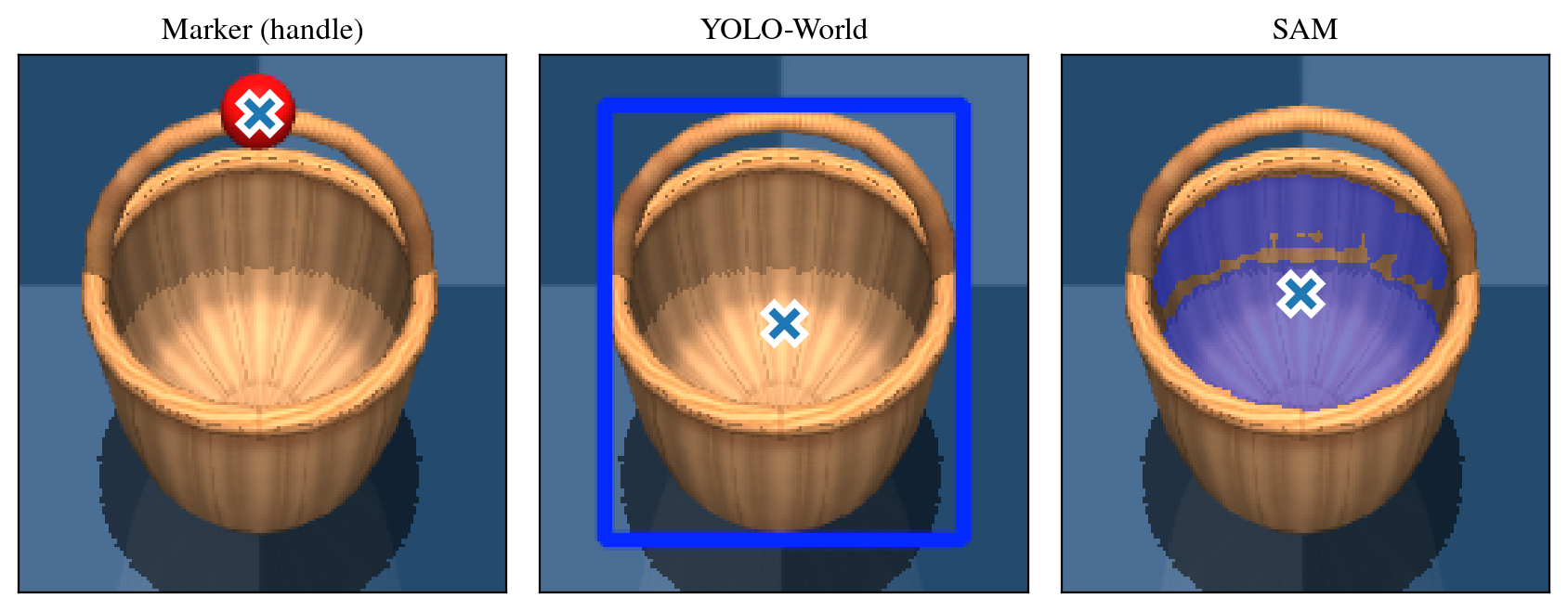}
    \caption{Marker against real YOLO-World and SAM outputs; the region centres (crosses) miss the handle.}
    \label{fig:perception}
\end{figure}

On the perception side, grasping relies on a fiducial marker, and a learned detector cannot yet replace it. YOLO-World~\cite{cheng2024yoloworld} and SAM~\cite{kirillov2023segment} localise the basket only as a box or mask (Figure~\ref{fig:perception}), with the region centre 13\,cm from the handle and well beyond the 4\,cm gripper aperture, so detection alone does not yield a usable grasp point $\mathbf{p}_g$. Turning such a region into a grasp pose needs a dedicated grasp-synthesis network that respects the arm's kinematics, which we leave for future work.

\section{Conclusion and Future Work}
\label{sec:conclusion}

Leg-integrated bimanual manipulators extend a quadruped into a loco-manipulation platform: the system picks up objects at ground level, coordinates both arms for two-handed tasks, and sequences these skills autonomously over a long horizon under a vision-language planner.

These results are obtained entirely in simulation. Building a physical prototype and reproducing the three tasks on real hardware is the immediate next step; it would test how the marker-based perception and the planted-stance grasping hold up under real sensing, actuation, and contact dynamics.

Beyond closing this sim-to-real gap, the same hardware admits further extensions. The prismatic slider and pitch joint open a second use beyond grasping: with $q_1$ extended and $q_2$ locked downward, each arm becomes a self-locking leg extension, raising the robot's stance height and enabling a stilt-like locomotion mode on the same hardware. Lifting a front leg would also let an arm reach back and set a grasped object onto the trunk, for instance into a basket on the back, freeing the arm for the next object. These are left to future work.

\bibliographystyle{IEEEtran}
\bibliography{refs}

\end{document}